\documentclass[preprint,5p,12pt]{elsarticle}

\usepackage[cmex10]{amsmath}

\usepackage{times}
\usepackage{graphicx,color,psfrag}
\usepackage{rotating}
\usepackage{algo,amsmath,amssymb,xspace}	

\usepackage{algo,amsmath,amssymb,fancyhdr,tabularx,xspace}

\usepackage{algorithmicx}
\usepackage[titletoc,toc,title]{appendix}

\renewcommand{\leq}{\leqslant}
\renewcommand{\geq}{\geqslant}

\usepackage{subfigure}

\journal{Expert Systems with Applications}

\begin{document}

\begin{frontmatter}
%
\title{Integrating K-means with Quadratic Programming Feature Selection}

\author[rvt]{Yamuna~Prasad\corref{cor1}\fnref{fn1}}
\ead{yprasad@cse.iitd.ac.in}
\ead[url]{http://www.cse.iitd.ernet.in/~yprasad}

\author[rvt]{K.~K.~Biswas\fnref{fn2}}
\ead{kkb@cse.iitd.ac.in}
\ead[url]{http://www.cse.iitd.ernet.in/~kkb}
%

\cortext[cor1]{Corresponding author}

\fntext[fn1]{Working as a research scholar in Department of CSE
 at Indian institute of Technology Delhi.Main areas of 
interests are machine learning, soft computing, optimization etc.}
\fntext[fn2]{Working as Professor in Department of CSE
 at Indian institute of Technology Delhi.Main areas of 
interests are AI, machine learning, soft computing, vision etc.}

\address[rvt]{Department of Computer Science and Engineering, Indian Institute of 
Technology Delhi, New Delhi, India \fnref{label3}}



\begin{abstract} 
Several data mining problems are characterized by data in high
dimensions. One of the popular ways to reduce the dimensionality
of the data is to perform feature selection, i.e, select a subset of 
relevant and non-redundant features. Recently, Quadratic Programming Feature
Selection (QPFS) has been proposed which formulates the feature selection problem
as a quadratic program. It has been shown to outperform many of the existing 
feature selection methods for a variety of applications. Though, better
than many existing approaches, the running time complexity of QPFS is 
cubic in the number of features, which can be quite computationally 
expensive even for moderately sized datasets.

In this paper we propose a novel method for feature selection by integrating 
k-means clustering with QPFS. The basic variant of our approach runs k-means to 
bring down the number of features which need to be passed on to QPFS. We then
enhance this idea, wherein we gradually refine the feature space from a very 
coarse clustering to a fine-grained one, by interleaving steps of QPFS with
k-means clustering. Every step of QPFS helps in identifying the clusters of 
irrelevant features (which can then be thrown away), whereas every step of 
k-means further refines the clusters which are potentially relevant. We show
that our iterative refinement of clusters is guaranteed to converge. We provide
bounds on the number of distance computations involved in the k-means algorithm. 
Further, each QPFS run is now cubic in number of clusters, which can be much 
smaller than actual number of features. Experiments on eight publicly available 
datasets show that our approach gives significant computational gains (both in
time and memory), over standard QPFS as well as other state of the art feature 
selection methods, even while improving the overall accuracy.
\end{abstract}

%

\begin{keyword}
Feature Selection \sep Support Vector Machine (SVM) \sep 
Quadratic Programming Feature Selection (QPFS)
\end{keyword}
\end{frontmatter}

\section{Introduction}
\label{intro}
Many data mining tasks are characterized by data in high dimensions. Directly 
dealing with such data leads to several problems including high computational costs and 
overfitting. Dimensionality reduction is used to deal with these problems by bringing down
the data to a lower dimensional space. For many scientific applications, each of the dimensions 
(features) have an inherent meaning and one needs to keep the original features (or a representative 
subset) around to perform any meaningful analysis on the data~\cite{cantu-paz&al04}. Hence, some of the 
standard dimensionality reduction techniques such as PCA which transform the original feature space can
not be directly applied. Dimensionality reduction in such scenarios reduces to the problem of feature
selection. The goal is to select a subset of features which are relevant and non-redundant. Searching 
for such an optimal subset is computationally intractable (search space is exponential)~\cite{Bins01,Yu08}. 
Amongst the current feature selection techniques, filter based methods are more popular because of the possibility 
of use with alternate classifiers and their reduced computational complexity (like Maximal relevance (MaxRel), 
Maximal Dependency (MaxDep), minimal-Redundancy-Maximal-Relevance (mRMR) 
etc.)~\cite{Bekkerman03,Forman03})~\cite{Lujan10}. 

Recently, a new filter based quadratic programming feature selection (QPFS) method~\cite{Lujan10} 
has been proposed which has been shown to outperform many other existing feature selection 
methods. 
In this approach, a similarity matrix representing the redundancy among the features and a 
feature relevance vector are computed. These together are fed into a quadratic program to get
a ranking on the features. The computation of the similarity matrix requires quadratic 
time and space in the number of features. Ranking requires cubic time in the number of features.
This cubic time complexity can be prohibitively expensive for carrying out feature selection
task in many datasets of practical interest. To deal with this problem, Lujan et al.~\cite{Lujan10} 
combine Nystr\"{o}m sampling method, which reduces the space and time requirement at the cost of 
accuracy. 

In this paper, we propose a feature selection approach by first clustering
the set of features using two-level k-means clustering~\cite{Chitta10} and then 
applying QPFS over the cluster representatives (called {\em Two-level K-Means QPFS}). 
The key intuition is to identify the redundant sets of features using k-means 
and use a single representative from each cluster for the ensuing QPFS run. This makes the 
feature selection task much more scalable since k-means has linear time complexity in the 
number of points to be clustered. The QPFS run is now cubic only in number of clusters, 
which typically is much smaller than actual number of features. Our approach is motivated by 
the work of Chitta and Murty~\cite{Chitta10}, which proposes a two-level k-means algorithm for 
clustering the set of similar data points and uses it for improving classification 
accuracy in SVMs. Chitta and Murty~\cite{Chitta10} show that their approach yields 
linear time complexity in contrast to standard cubic time complexity for SVM training. 

We further enhance our feature selection approach by realizing that instead
of simply doing one pass of k-means followed by QPFS, we can run them
repeatedly to get better feature clusters. Specifically, we propose 
a novel method for feature selection by interleaving steps of QPFS with 
MacQueen's~\cite{MacQueen67} k-means clustering (called {\em Interleaved K-means
QPFS}). We gradually refine the feature space from a very coarse clustering to a 
fine-grained one. While every step of QPFS helps in identifying the clusters of 
irrelevant features (i.e. having $0$ weights for the representative features), every step of k-means 
refines the potentially relevant clusters. Clusters of irrelevant features are thrown 
away after every QPFS step reducing the time requirements. This process is repeated 
recursively for a fixed number of levels or until each cluster has sufficiently small 
radius. Each QPFS run is now cubic in the number of clusters (which are much smaller than actual 
number of features and may be assumed to be constant). We show that our algorithm is guaranteed
to converge. Further, we can bound the number of distance computations employed during the k-means 
algorithm. 

We perform extensive evaluation of our proposed approach on eight publicly available 
benchmark datasets. We compare the performance with standard QPFS as well as other 
state of the art feature selection methods. Our experiments show that our approach 
gives significant computational gains (both in time and memory), even while improving the 
overall accuracy. 

In addition to Chitta and Murty~\cite{Chitta10}, there is other prior literature which 
uses clustering to reduce the dimensionality of the data for classification and related tasks. 
Examples include Clustering based SVM (CB-SVM)~\cite{yu&al03}, clustering based trees for k-nearest 
neighbor classification~\cite{zhang04} and use of PCA for efficient Gaussian kernel summation~\cite{lee&gray08}. 
\cite{liu05} presents a framework for categorizing existing feature selection
algorithms and chosing the right algorithm for an application based on data characteristics. To the 
best of our knowledge, ours is the first work which integrates the use of clustering with existing 
feature selection methods to boost up their performance. Unlike most previous approaches, which use 
clustering as a one pass algorithm, our work interleaves steps of clustering with feature selection,
thereby, reaping the advantage of clustering at various levels of granularity. The key contributions of 
our work can be summarized as follows:
\begin{itemize}
\item A novel way to integrate the use of clustering (k-means) with existing feature selection methods (QPFS)
\item Bounds on the performance of the proposed algorithm
\item An extensive evaluation on eight different publicly available datasets
\end{itemize}

The rest of the paper is organized as follows: We describe the background for QPFS approach and 
the two level k-means algorithm in Section~\ref{back}. Our proposed Two-level k-means QPFS and 
Interleaved K-Means QPFS approaches are presented in Sections~\ref{varkm} and~\ref{ikmqpfs}, 
respectively. Experimental results are described in Section~\ref{expe}. We conclude our work 
in Section~\ref{conc}.
%


\section{Background}
\label{back}

%
%

\subsection{QPFS~\cite{Lujan10}}
\label{qpfs}
Given a dataset with $M$ features ($f_i, i = 1, ..., M$) and $N$ training instances ($x_i, i= 1, ..., N$) 
with Class $Y$ labels ($y_i, i= 1, ..., Y$) the standard QPFS formulation~\cite{Lujan10} is:
\begin{equation}
\label{eq1}
\begin{aligned}
& & f\left(\alpha\right) & = \min_{\alpha} \frac{1}{2}\alpha^TQ\alpha - s^T\alpha \\
& \text{Subject to}
& & \alpha_i \geq 0,   i = 1, ..., M; \qquad I^T\alpha = 1.
\end{aligned}
\end{equation} 
where, $\alpha$ is an $M$ dimensional vector, $I$ is the vector of all ones and $Q$ is an $M\times M$ 
symmetric positive semi-definite matrix, which represents
the redundancy among the features;
$s$ is an $M$ size vector representing relevance score of features with respective class labels.
In this formulation, the quadratic term captures the dependence between each pair of features,
and the linear term captures the relevance between each of the features and the class labels.
The task of feature selection involves optimizing the twin goal of selecting features with high
 relevance and low redundancy.
Considering the relative importance of non-redundancy amongst the features 
and their relevance,a scalar quantity
$\theta \in \left[0, 1\right]$ is introduced in the above formulation resulting in~\cite{Lujan10}: 
\begin{equation}
\label{eq2}
\begin{aligned}
& & f\left(\alpha\right) & = \min_{\alpha} \frac{1}{2}\left(1-\theta\right)\alpha^TQ\alpha - \theta s^T\alpha \\
& \text{Subject to}
& & \alpha_i \geq 0, i = 1, ..., M; \qquad I^T\alpha = 1.
\end{aligned}
\end{equation} 
In the above equation, $\theta = 1$ corresponds to the formulation where only relevance is considered. 
In this case the QPFS formulation becomes equivalent to Maximum relevance criterion.
 When $\theta$ is set to zero, the formulation considers only non-redundancy among the features, that is, 
features with low redundancy with the rest of the features are likely to be selected. 
A reasonable value of $\theta$ can be computed using
\begin{equation*}
\label{eq2a}
\begin{aligned}
& & \theta = \bar{q} / (\bar{q} + \bar{m}) \\
\end{aligned}
\tag{2a}
\end{equation*}
where, $\bar{q}$ is the mean value of the elements of matrix $Q$ and $\bar{m}$ is the mean value of the 
elements of vector $s$.
As $\theta$ is a scalar, the similarity matrix $Q$ and the feature relevance vector $s$ in~\eqref{eq2} 
can be scaled according to the value of $\theta$, resulting in the equivalent QPFS formulation of Equation~\eqref{eq1}.
The QPFS can be solved by using any of the standard quadratic programming implementations 
but it raises space and computational time issues. Time complexity of QPFS approach 
is $O(M^3 + NM^2)$ and space complexity is $O(M^2)$.
To handle large scale data, Lujan et al.~\cite{Lujan10} proposes to combine QPFS with Nystr\"{o}m 
method by working on subsamples of the data set for faster convergence. This often comes at the cost
of trade-off with accuracy. The details are available in~\cite{Lujan10}.

\subsection{Similarity Measure}
\label{sm}
Various measures have been employed to represent similarities among features~\cite{Au05,Yu08,Jornsten03,Bins01}. 
Among these, correlation and mutual information (MI) based 
similarity measures are more popular. The classification accuracy can be improved with MI as it captures nonlinear 
dependencies between pair of variables unlike correlation coefficient which only measures linear 
relationship between a pair of variables~\cite{Peng05,Lujan10}. The mutual information between a 
pair of features $f_i$ and $f_j$ can be computed as follows:
\begin{equation}
\label{smeq}
MI(f_i,f_j) = H(f_i) + H(f_2) - H(f_i,f_2) 
\end{equation}
where $H(f_i)$ reperents entropy of feature vector $f_i$ and $H(f_i,f_j)$ represents
the joint entropy between feature vectors $f_i$ and $f_j$~\cite{Yao03}. 
Following variant of mutual information can be used as distance metric~\cite{Yao03}:
\begin{equation}
\label{eqnmi}
 d(f_i,f_j) = 1 - \frac{MI(f_i,f_j)}{\max (H(f_i),H(f_j))}
\end{equation}


\subsection{MacQueen's K-Means Algorithm~\cite{MacQueen67}}
\label{mk}
This is a k-means clustering algorithm which runs in two passes. In the first pass, it 
chooses first $k$ samples as the initial $k$ centers and assigns each of the remaining 
$N-k$ samples to the cluster whose center is nearest and updates the centers. In the second 
pass, each of the $N$ samples is assigned to the clusters whose center is closest and centers 
are updated. The number of distance computations in the first and second passes are  $k(N-k)$ 
and $Nk$ respectively. Thus, the number of distance computations needed in MacQueen's k-means 
algorithm is $2Nk - k^2$. This in effect means that the complexity is $O(Nk)$~\cite{Chitta10}.

\subsection{Two-level K-means\cite{Chitta10} Algorithm}
\label{tlk}
Recently, a two-level k-means algorithm has been developed using MacQueen's k-means 
algorithm~\cite{Chitta10}. This clustering algorithm ensures that radii of the clusters 
produced is less than a pre-defined threshold $\tau$. The algorithm is outlined below:
\begin{algorithm}{Two-level K-means}[D, k,\tau]{\qinput{Data Set $D$, Initial Number of 
Clusters $k$ and Radius Threshold $\tau$.}
 \qoutput{Set of clusters $C$ ($c_1, c_2, \ldots,c_i, \ldots$) (with radius 
 $r_i \leq \tau$) and the set of cluster centers $\mu$.}
 \label{im}
 }
  ({\bf level 1: }) Cluster the given set of data points into an arbitarily 
 chosen $k'$ clusters using MacQueen's k-means algorithm.
 \\ Calculate the radius $r_i$ of i$^{th}$ cluster using $r_i = \max_{x_j \in c_i} d(x_j,c_i)$, 
where, $d(. , .)$ is the similarity metric. 
  \\ ({\bf level 2: }) If the radius $r_i$ of the cluster $c_i$ is greater than the user 
  defined threshold $\tau$, split it using MacQueen's k-means with the number of 
  clusters set to $(\frac{r_i}{\tau})^M$, where $M$ is the dimension of the data. 
        \\	\qreturn the set of clusters ($C$) and corresponding centers ($\mu$)
		obtained after level 2.
        \qend
\end{algorithm}

~\cite{Chitta10} shows that the above two-level k-means algorithm reduces the number of distance 
calculations as required by the MacQueen's k-means algorithm, while guaranteeing a bound on the 
clustering error (details below). 
The difference between the number of distance computations by MacQueen's k-means algorithm 
and the two-level k-means algorithm follows the inequality:
%
\begin{equation}
\label{eq3}
  U - \frac{N\alpha^MR}{\tau} \leq ND_1 - ND_{2} \leq U + \frac{k'\alpha^{2M}R}{2\tau}
\end{equation}
where, $ND_1$ and $ND_2$ are the distance computations in MacQueen's k-means algorithm and two-level 
k-means algorithm respectively,$\alpha \geq 0$ is some constant, $R$ is the radius of the ball 
enclosing all the data points and $U$ is $(k-k')(2N-k-k')$.
If $k' \ll k$, then the expected number of distance computations in level 2 is upper bounded
 by $N\alpha^MR/\tau$. The parameter $\tau$ obeys the following inequality:
 \begin{equation*}
  \frac{N\alpha^MR}{U} \leq \tau \leq R
 \end{equation*}
An appropriate choice of $\tau$ is obtained using the inequality
\begin{equation}
\label{eq4}
 \max \big(\frac{R}{(k)^{1/M}}, \frac{R}{(2N-k)^{1/M}}\big) \leq \tau \leq R
\end{equation}
The clustering error in two-level k-means algorithm is upper bounded by 
twice the error of optimal clustering~\cite{Chitta10}. The time complexity of 
two-level k-means algorithm is $O(Nk)$ and the space complexity is $O(N+k)$~\cite{Chitta10}. 
The detailed analysis of these bounds can be found in~\cite{Chitta10}.

\section{Two-level K-means QPFS}
\label{varkm}
%
%
%

Authors in~\cite{Chitta10} employ two-level k-means clustering for reducing
the number of data points for classification using SVM.
We use similar idea except that we cluster a set of features instead of the set of data
points.
%
%
We then apply QPFS on representative set of features. 
Thus, the problem is transformed into the feature space in contrast with their formultaion
in the space of data points. Another key distinction is that we 
need to work with actual features unlike cluster means as in the case of~\cite{Chitta10}.
This is because the means of feature clusters are abstract points and may not 
correspond to an actual features over which feature selection could be 
carried out. Towards this end, we develop two algorithms, the first one by modifying 
the MacQueen's k-means algorithm and the other one by modifying the two-level k-means 
algorithm~\cite{Chitta10} to return cluster representatives (features) in place of cluster 
means. Each feature is represented as an $N$-dimensional vector where $N$ denotes
the number of training instances (see Section~\ref{qpfs}). The $k^{th}$ component of this 
vector denotes the value of the feature in the $k^{th}$ data point. The distance metric 
between a pair features is defined using mutual information as in Equation~\eqref{eqnmi}.

In the following sections, $M$ is cardinality of the (feature) space to be clustered. 
This takes the place of $N$ which is the cardinality of (data) space in the case of
Chitta and Murty~\cite{Chitta10}. Similarly, $N$ denotes the dimensionality of the (feature) 
space to be clustered. This takes place of $n$ which is the dimensionality of the (data) 
space in case of Chitta and Murty~\cite{Chitta10}.

%

\subsection{Variant MacQueen's K-means}
\label{vmkm}
We propose a variant of MacQueen's K-means algorithm for clustering the features to produce
set of clusters with redundant features instead of clustering datapoints. 
In each iteration of the MacQueen's K-means algorithm,
the nearest point from the updated mean is selected as the new center
(called the {\em cluster representative}).
Each iteration needs to compute distance from $M-k$ features to $k$ centers and distance from center to 
nearest feature in its cluster. Thus, each iteration needs $k(M-k)+M$ distance computations.
As MacQueen's k-means uses two iterations, the total number of distance compuations 
would be $2Mk - 2k^2 + 2M$. The complexity is thus $O(Mk)$.
%

\subsection{Variant Two-level K-Means(TLKM)}
\label{tlkm}
We propose two-level k-Means algorithm (TLKM) by replacing MacQueen's k-means algorithm with
its variant in the two-level k-Means algorithm as given in Section~\ref{tlk}. 
It is important to note that we are clustering features rather than the data points.
TLKM returns the feature clusters along with corresponding representatives. Following the 
arguments in~\cite{Chitta10}, we can derive the bounds on number of distance computations for 
our proposed TLKM algorithm in a similar manner. The only difference is that we have an additional 
$2M - k^2$ term as explained in Section~\ref{tlk} 
The bounds for difference in the number of distance computations between variant MacQueen's k-means 
and TLKM is
\begin{equation}
 \label{eq5}
  U - M(\frac{(\alpha^N+1)R}{\tau}) \leq ND_1 - ND_{2} \leq U + \frac{k'\alpha^{2N}R}{\tau}  
\end{equation}
Here, $U$ is $2(k-k')(M-k-k')$ and other parameters have same definitions as in Equation~\eqref{eq3}
of Section~\ref{tlk}.
Further, if $k' \ll k$, then the expected number of distance computations in the second level is upper bounded
 by $M(2+ (\alpha^N+1)R/\tau)$ and parameter $\tau$ obeys the following inequality
 \begin{equation*}
  M\big(\frac{(\alpha^N+1)R}{U}\big) \leq \tau \leq R
 \end{equation*}
 Following~\cite{Chitta10}, for reducing the number of computations 
in TLKM algoritm, it is necessary that
\begin{equation}
\label{eq6}
 \max \big(\frac{R}{(k)^{1/N}}, \frac{R}{(M-k)^{1/N}}\big) \leq \tau \leq R
\end{equation}
OR,
\begin{equation}
\label{eq7}
  \tau \leq \min \big(\frac{R}{(k)^{1/N}}, \frac{R}{(M-k)^{1/N}}\big)\leq R
\end{equation}
Following the arguments in~\cite{Chitta10}, it can be shown that the time and space 
complexities of the modified two-level k-means for clustering features are $O(Mk)$ and 
$O(M+k)$, respectively.

\subsection{Two-level K-Means QPFS (TLKM-QPFS) Algorithm}
We are now ready to present the QPFS based feature selection method using TLKM.
We named this algorithm TLKM-QPFS, henceforth. We employ TLKM approach to 
cluster the features in a given dataset followed by a run of QPFS. 
Algorithm~\ref{tlkm-qpfs} illustrates our proposed 
Two-level k-means QPFS (TLKM-QPFS) approach.
\begin{algorithm}{TLKM-QPFS}[FS, k,\tau]{\qinput{Feature Set $FS$, Initial 
Number of Clusters $k$ and
 Radius Threshold $\tau$.}
 \qoutput{Final representative feature set $F$ (features) in order of their $\alpha$ values.}
 \label{tlkm-qpfs}
 }
  Find the representatives $F$ using TLKM algorithm as defined in Section~\ref{tlkm} \label{vts1}
 \\ Apply QPFS on the cluster representatives $F$. \label{vts2}
        \\	\qreturn Ranked $F$ in the order of $\alpha$
        \qend
\end{algorithm}
Time and space complexities for TLKM approach in Step~\ref{vts1} are $O(Mk)$ and $O(M+k)$ 
respectively. In step~\ref{vts2} of Algorithm~\ref{tlkm-qpfs}, QPFS approach is used to rank the $k$ cluster representatives 
(features) obtained in step~\ref{vts1}. Time and space complexities for this step are 
$O(k^3 + Nk^2)$ and $O(k^2)$, respectively. Therefore, the total time and space complexities of 
the algorithm~\ref{tlkm-qpfs} are $O(Mk) +O(k^3 + Nk^2) \sim O(M)$ and $O(M+k)+O(k^2) \sim O(M)$,
respectively. It is clear from this analysis that both the time and space complexities of this 
algorithm are $O(M)$ as $k \ll M$.

\section{Interleaved K-Means QPFS (IKM-QPFS)}
\label{ikmqpfs}

We now propose a new algorithm by combining the benefits of clustering approach with QPFS.
In this proposed algorithm, we strive to refine relevant feature space from coarse to 
fine-grained clusters to improve accuracy while still preserving some of the computational 
gains obtained by TLKM-QPFS. Algorithm~\ref{tlkm-qpfs} uses k-means to identify cluster of features 
which are similar to each other (redundant). A representative is chosen for each of the clusters 
and then fed into QPFS. QPFS in turn returns a ranking on these cluster representatives. 
Many of the representatives are deemed irrelevant for classification ($\alpha =0$).
Amongst the sets of clusters whose representatives were deemed irrelevant, consider those 
with cluster radius $r < \tau$. All the features in the these clusters can be considered 
irrelevant (since the cluster representative was irrelevant and cluster radius is sufficiently 
small) and can be thrown away. This also gives us an opportunity to further refine the larger 
clusters($r>\tau$) potentially improving accuracy by identifying a larger subset of relevant 
features.  This process of executing QPFS after initial run of k-means clustering can be repeated 
recursively. Each run of k-means further refines the relevant sub-clusters whereas 
each run of QPFS helps in identifying  relevant set of features. This leads 
to the following algorithm for feature selection which we have named Interleaved K-Means 
QPFS (IKM-QPFS).

\subsection{Interleaved K-Means QPFS (IKM-QPFS) Algorithm}
To start with, we first employ k-means to find the a set of cluster representatives.
These cluster representatives are then fed into QPFS to get feature
ranking on them. The cluster with sufficiently small radius ($r<\tau$)
need not be refined further and can be directly use for final level of 
feature selection. Here, we throw away those representatives whose
$\alpha$ values are zero(irrelevant for classification). At the 
same time, clusters with radius greater than $\tau$ need to be refined
further. This can be done recursively using above steps.
In practice, we need to run the recursive splitting of clusters 
only upto a user defined level. In our approach, we split each
cluster into a fixed number ($k$) of sub-clusters during k-means
splitting.
The proposed Interleaved K-Means-QPFS algorithm is presented in Algorithm~\ref{im}.
\begin{algorithm}{IKM-QPFS}[FS, k, L, \tau]{\qinput{Feature Set $FS$, Number of Sub-Clusters 
$k$ that each Cluster is split into, Radius Threshold $\tau$ and  Number of Interleaved Levels 
$L$.}
\qoutput{Ordered set of relevant features $F$}
 \label{im}
 }
  	    Apply variant MacQueen's algorithm to features in $FS$; Obtain clusters $C$, cluster representatives $f$. \label{ims1}
  \\	Apply QPFS on the cluster representatives $f$ and obtain $\alpha$.\label{ims2}
  \\    $l \qlet 1$
  \\	$F \qlet \qproc{IRR}(C,f,\alpha,k,\tau,l,L)$ \label{ims3}
  \\	Apply QPFS on $F$ and rank $F$ according to $\alpha$. \label{ims4}
        \\	\qreturn $F$
        \qend
\end{algorithm}
The sub procedure IRR(Identify Relevant and Refine) is illustrated in Algorithm~\ref{irr}.
\begin{algorithm}{IRR}[C,f,\alpha,k,\tau,l,L]{\qinput{Cluster Set $C$, Cluster 
 representatives $f$, $\alpha$ obtained by QPFS, Number of Clusters $k$, 
Radius Threshold $\tau$, Number of level $l$, and 
Maximum Number of Levels $L$}
 \qoutput{Final centers $F$ (features) in order of their $\alpha$ values.}
 \label{irr}
 }
  \qfor each cluster $c_i \in C$ 
  \\   \qdo 
  \\      find the radius $r_i = \max_{f_j \in c_i} d(f_j,f_i)$; $d(.,.)$ is the distance metric. 
  \\	  \qif ($r_i < \tau$ or $l = L$) \label{ccrs1}
  \\		\qthen \qif($\alpha_i > 0$)
  \\			     \qthen $F = \cup \lbrace f_i \rbrace$ \label{ccrs2}
                   \qfi
  \\       \qelse \label{ccrs3}
  \\				Apply variant MacQueen's k-means algorithm to features in cluster $c_i$; Obtain clusters $C'$, cluster representatives $f'$\label{ccrs4}
  \\				Apply QPFS on the cluster centers $C'$ and get $\alpha'$. \label{ccrs5}
  \\				$l \qlet l+1$
  \\				$F'\qlet\qproc{IRR}(C',f',\alpha',k,\tau,l)$ \label{ccrs6}
  \\				$F \qlet F \cup \lbrace F' \rbrace$
	\qfi
	\qend
        \\	\qreturn $F$
        \qend
\end{algorithm}
In the above algorithm, {\em if} condition in step~\ref{ccrs1} checks if
the boundary condition has been reached and no more splitting needs to be done 
(i.e. maximum number of levels $L$ has been reached or $r_i < \tau$). In which case, 
if the cluster is relevant ($\alpha_i > 0$), then corresponding features are added to 
the feature set to be returned (step~\ref{ccrs2}). Else, they are discarded. {\em Else} 
condition in step~\ref{ccrs3} goes on to recursively refine the clusters when boundary 
condition is not yet reached.

The recursive approach for a sub-cluster at i$^{th}$ level can be visualized as follows.
\begin{figure}[!htb]
 \centering\input{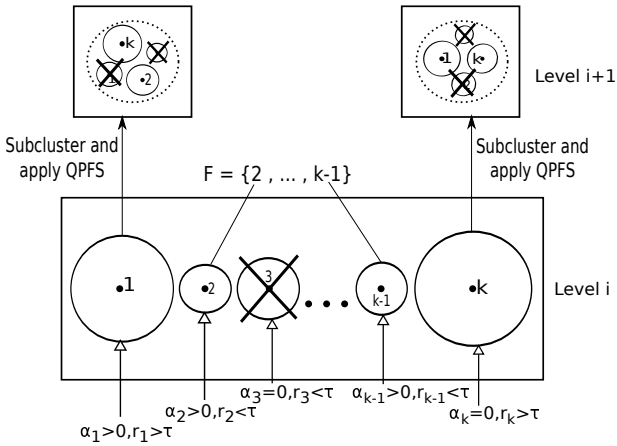}
 \caption{Sub-clustering process at level $i$}
 \label{figuai}
\end{figure}
 In Figure~\ref{figuai}, sub-clusters $1$ and $k$ have radii greater
than $\tau$. They are split further independent of $\alpha$ values.
Their contribution to the final feature set is calculated by refining
them recursively. Sub-clusters $2$, $3$ and $k-1$ have radii less than 
$\tau$. They don't need to be split further. Amongst these,
representatives for $2$ and $k-1$ contribute to the final set of features. 
Sub-cluster $3$ is discarded since $\alpha_3 =0$.  

\subsection{Convergence}
In every recursive call of Algorithm~\ref{irr}, all the clusters whose
radius is greater than $\tau$ are further split into $k$ sub-clusters. 
Since every split is guaranteed to decrease the size of the original
cluster, and we have a finite number of features, the algorithm
is guaranteed to terminate and find clusters each of whose radius
is less than $\tau$, given sufficiently large $L$. Note that in
the extreme case, a cluster will have only one point in it and 
hence, its radius will be zero. Now, let us try to analyze what
happens in an average case i.e. when the sub-cluster split induced 
by the MacQueen's algorithm results in uniform-sized clusters. More 
formally, let $r_i$ denote the radius of the cluster $i$ (at some 
level) which needs to be split further. Then, the volume enclosed
by this cluster is $C*{r_i}^N$. Here, $N$ is the number of original
data points (this is the space in which features are embedded). 
By the assumption of uniform size, this volume is divided equally amongst 
all the sub-clusters. Hence, the volume of each sub-cluster is going to be
$C*{r_i}^N/k$. This volume corresponds to a sub-cluster of radius $r_i/k^{1/N}$. 
Hence, at every level, the cluster radius is reduced by a factor of $k^{1/N}$. If 
the starting radius is $R$, then, after $l$ levels the radius of a sub-cluster is 
given by $R/k^{l/{N}}$. We would like this quantity to be less than equal to $\tau$. 
This results in the following bound on $l$. 
\begin{align*}
\frac{R}{k^{\frac{l}{N}}} \le \tau \quad\implies \quad k^{\frac{l}{N}} \ge \frac{R}{\tau} \quad \implies \quad
k^l \ge \left(\frac{R}{\tau} \right)^{N} \\
\quad \implies l \quad\ge N*\log_k \left(\frac{R}{\tau}\right) \text{ (taking log)}\qquad \qquad \qquad \quad 
\end{align*}
Hence, under the assumption of uniform splitting, continuation
up to $N*\log_k(R/\tau)$ levels will guarantee that each sub-cluster 
has radius $\le \tau$. If $N \approx M$, then features are very 
sparsely distributed in the data space, and above is a very loose
bound. On the other hand if $N \ll M$ (as is the case with many Microarray
datasets), then, above bound can be put to practical use. 

MacQueen's algorithm starts with the first set of $k$ points as the cluster
representatives, followed by another pass of assigning the points to each cluster 
and then recalculating the cluster representatives. In general, the assumption of
uniform sub-cluster may only be an approximation to the actual clusters which are 
obtained, and hence, above bound will also be an approximation. A detailed analysis 
of whether one can bound this approximation is proposed to be carried out in future.

\subsection{Distance Computations}
Distance computations done by interleaved steps of k-means in 
Algorithm~\ref{ikmqpfs} can be bounded as follows. In worst
case, none of the clusters will be discarded and also, their
radii will be greater than or equal to threshold ($\tau$) at each
level. This will lead to recursive splitting of each cluster upto
level $L$. Now, consider cluster $c_j$ at level $i$. The number
of distance computations required by the MacQueen's algorithm
to split this cluster further is given by $2|c_j|k-2k^2+2|c_j|$
(see Section~\ref{vmkm}). Thus, total number of distance computations
at level $i$ is given as $\Sigma_{j} 2|c_j|k-2k^2+2|c_j| = 2Mk-k^2+2M$.
The equality follows from the fact that total number of points
in clusters at any level is $\Sigma_j |c_j|=M$ (since each cluster is 
split upto the last level). Therefore, the number of distance computations
in worst case is independent of the particular level. Hence, the total
number of distance computations for Algorithm~\ref{ikmqpfs} can be 
bounded by $L(2Mk-k^2+2M)$.

\subsection{Time Complexity Analysis}
Time required in step~\ref{ims1} and step~\ref{ims2} of Algorithm~\ref{im} is $O(Mk)$ 
and $O(k^3)$ respectively. In step~\ref{ims3} of Algorithm~\ref{im}, Algorithm~\ref{irr} 
is called which is executed recursively. The time required for its execution can be computed 
as follows:\\ 
If the maximum number of levels is $L$, number of cluster is $k$ then it can be easily 
shown that the time complexity of Algorithm~\ref{irr}
is upper bounded by $O(LMk + k^{3+L})$. The first term comes from the number
distance computations in k-means, and the second term comes from the $O(k^L)$
calls to QPFS ($k^{i-1}$ calls at level $i$), where each call takes $O(k^3)$ time.\\ 
Thus, time required in step~\ref{ims3} of Algorithm~\ref{im} is $O(LMk + k^{3+L})$ and 
time required in step~\ref{ims4} of Algorithm~\ref{im} is $O(k^{L+3})$. 

As $L$ and $k$ are very small constants, total time required by  
Algorithm~\ref{im} is $O(Mk) + O(k^3) + O(LMk + k^{3+L}) + O(k^{L+3})$ 
which in effect is $O(M)$.

\subsection{Interleaved K-Means Aggressive QPFS (IKMA-QPFS)}
In this section, we present a variation on the IKM-QPFS algorithm
described above. The key idea is that after every step of QPFS
run, we throw away all the clusters whose representatives are
deemed irrelevant during a QPFS run (i.e., $\alpha=0$), indepedent of 
the radii of the corresponding clusters. This is a deviation from the 
original proposed algorithm, wherein, we throw away a cluster only if the 
corresponding $\alpha=0$ and the cluster radius $r \le \tau$. We call this variation 
Interleaved K-Means Aggressive QPFS (IKMA-QPFS) since it is aggressive 
about discarding the clusters whose representatives are deemed irrelevant. 
This potentially leads to even larger gain in terms of computational complexity
since IKMA-QPFS tries to identify the irrelvant feature clusters early enough
in the process and throws them away. But since some of these clusters can be large 
in size ($r \ge \tau$), we might trade-off the additional computational gain by a 
loss in accuracy. But interestingly, in our analysis, we found that almost 
always this aggressive throwing away of clusters happened only towards the deeper levels
of clustering(i.e., very few representatives were deemed irrelevant in the beginning
levels of clustering), where the clusters were already sufficiently small. Hence,
as we will see in our experiments, not only this variant performs better
in terms of computational efficiency than IKM-QPFS, it even simplifies the 
feature selection problem, giving improved accuracy in some cases.

For IKMA-QPFS, the only change in the Algorithm~\ref{irr} is before step 3 (i.e., 
right after the {\tt for} loop starts), where we need to put another check 
{\tt if($\alpha_i = 0$)}. If this condition is satisfied, 
we simply return out of the function. Rest of the algorithm remains the same. 
The convergence, the distance computations and the time complexity analyses presented 
above also remain the same as for IKM-QPFS. This is because all the analyses have been 
done in the worst case when no clusters might be thrown away at intermediate levels.

\section{Experiments}
\label{expe}
%

We compare the performance of our proposed approaches TLKM-QPFS,
IKM-QPFS and IKMA-QPFS with QPFS, FGM and GDM on eight publicly 
available benchmark datasets.
We compare all methods for their time and memory
requirements and also for their error rates at various numbers of
top-k features selected. FGM and GDM methods works for binary classification datasets, 
therefore comparison with FGM and GDM is not carried out for SRBCT multi-class 
classification datasets. We observe an improved accuracy for FGM and GDM on normalized 
dataset in range $\lbrack$ -1, 1$\rbrack$. Therefore, we normalized all the datasets in
range $\lbrack$ -1, 1$\rbrack$.

We plot the accuracy graphs for varying ($1$ to $100$) the number of top-k 
features selected for all the datasets except WDBC. For WDBC dataset, we have reported
the results up-till 30 top features as this dataset has only 30 features. Next we describe 
the details of the datasets and our experimental methodology followed by our actual results.
\subsection{Datasets}
\label{dataset}
For our experimental study, we have used eight publicly available
benchmark datasets used by other researchers for feature selection. 
The description of these datasets is presented in 
Table~\ref{table1}. WDBC is breast cancer Wisconsin (diagnostic)  dataset, 
Colon, SRBCT, Lymphoma, Leukemia and RAC datasets are microarray datasets
(~\cite{Lujan10,Ganesh12,Yu08,mingkui10}) and the last two are 
vision~\cite{mingkui10,yiteng12} datasets. 
\setlength{\tabcolsep}{7pt}
\begin{table}[!htb]
\begin{center}
{\caption{Datasets: detailed description}\label{table1}}
\begin{tabular}{lrrcr}
\hline
\rule{0pt}{12pt}
 & {\bf No. of }&{\bf No. of }&{\bf No. of }& \\
{\bf Dataset} & {\bf Instances }&{\bf Features}&{\bf Classes}
\\
\hline
\\[-6pt]
\quad WDBC&569&30&2 \\ 
\quad Colon&62&2000&2 \\ 
\quad SRBCT&63&2308&4 \\ 
\quad Lymphoma&45&4026&2 \\ 
\quad Leukemia&72&7129&2 \\ 
\quad RAC&33&48701&2 \\ 
\quad MNIST&13966&784&2 \\ 
\quad USPS&1500&241&2 
\\
\hline
\end{tabular}
\end{center}
\end{table}
\subsection{Methodology}
\label{methodology}
WDBC and USPS datasets are divided into 60\% and 40\% sized splits for training 
and testing, respectively as in~\cite{mingkui10}. MNIST dataset is divided in 11,982 training
and 1984 testing instances following~\cite{yiteng12}.
The reported results are the average over 100 random splits of the data. 
The number of samples is very small (less than 100) in microarray datasets,
 so leave-one-out cross-validation is used for these datasets. 
We use mutual information as in ~\cite{Lujan10} for redundancy and relevance measures in the 
experiments. The data is discretized using three 
segments and one standard deviation for computing mutual information as in~\cite{Lujan10}. 
For QPFS, the value of scale parameter ($\theta$) is computed using cross-validation from the 
set of $\theta$ values $\{$0.0, 0.1, 0.3, 0.5, 0.7, 0.9$\}$. 
The error rates obtained were very similar to the ones obtained using the scale parameter based 
on Equation~\eqref{eq2a}.
For TLKM-QPFS, we used cross validation to determine the values of expected number of clusters $k$
(to get $\tau$) and for IKM-QPFS and IKMA-QPFS, we used cross validation to determine the good values 
$\tau$ (threshold parameter) and $k'$ (initial number of clusters). Threshold parameter $\tau$ is choosen from
the set $\{$0.70, $\ldots$,0.99$\}$ with step size of 0.01. $k$ is choosen from the set $\lbrace$ 5,...,1000
 $\rbrace$ with step size of 5 and $k'$ parameter in IKM-QPFS (as well in IKMA-QPFS) is choosen from the 
set $\lbrack$ 3, 150 $\rbrack$.
%
%

After feature selection is done, linear SVM (L2-regularized L2-loss support vector 
classification in primal)~\cite{Fan08} is used to train a classifier using the
optimal set of features output by QPFS, TLKM-QPFS, IKM-QPFS and IKMA-QPFS methods. 
FGM and GDM are embedded methods, so accuracy for both of these methods are
 obtained according to~\cite{mingkui10,yiteng12}.
The experiments were run on a Intel Core$^{TM}$ i7 (3.10 GHz) machine with 8 GB RAM. 

\begin{figure}
 \begin{center}
\input{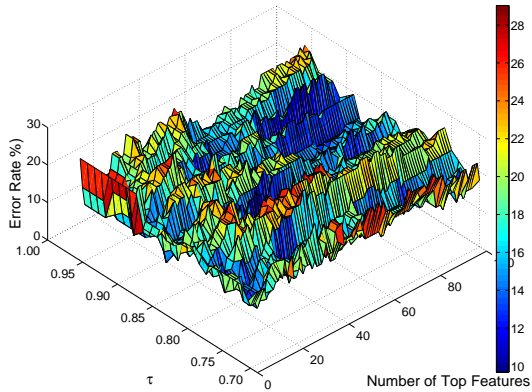}
 \end{center}
 \caption{Plot of accuracies (\%) for Colon dataset using IKM-QPFS with varying $\tau$ and varying number 
of top $k$(1-100) features at fixed initial clusters $k'$=15}
\label{fig4}
\end{figure}

\begin{figure}
 \begin{center}
\input{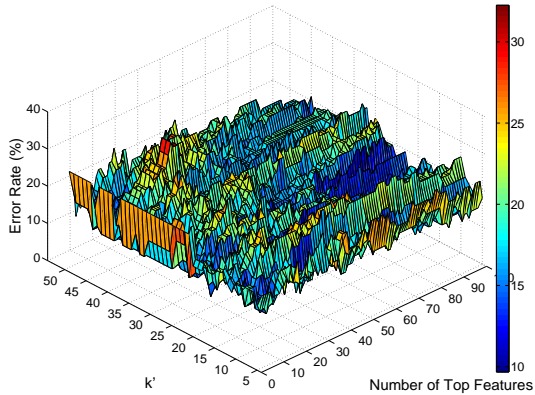}
 \end{center}
 \caption{Plot of accuracies (\%) for Colon dataset using IKM-QPFS with varying number of initial clusters
 $k'$ and varying number of top $k$(1-100) features at fixed $\tau$=0.8}
\label{fig5}
\end{figure}

We have presented the variation in the error rates on varying the values of $\tau$ with a
fixed value of initial number of clusters $k'$=15 in figure~\ref{fig4} and on varying the
values of initial number of clusters $k'$ with a fixed value of $\tau$=0.8 in figure
~\ref{fig5} for Colon dataset. On other datasets, it shows a similar trend.
\subsection{Results}
\label{results}
\subsubsection{Time and Memory}
\label{tmr}
Tables~\ref{table2-1} and~\ref{table2-2} show the time and memory 
requirements for feature selection done
using each of the methods for all datasets respectively. On all the 
datasets, TLKM-QPFS,IKM-QPFS 
and IKMA-QPFS are orders of magnitude faster than QPFS.
TLKM-QPFS is three times faster than the GDM on RAC, MNIST and USPS datasets while 
three to ten times slower than the GDM on WDBC, Colon, Lymphoma and Leukemia datasets.
Further, TLKM-QPFS is an order of magnitude faster than the FGM on MNIST and USPS
 datasets. IKM-QPFS and IKMA-QPFS are three to five times faster than FGM and GDM on 
RAC, MNIST and USPS datasets while two to five times slower on WDBC, Colon, Lymphoma
 and Leukemia datasets.
 The performance of TLKM-QPFS and IKM-QPFS are comparable while IKMA-QPFS is two to
fifteen times faster than TLKM-QPFS and two to six times faster than the IKM-QPFS. This 
achievement of reduction in time of IKMA-QPFS is due to aggressive throwing of 
clusters when $\alpha$ becomes zero.
%
%

QPFS ran out of memory for RAC dataset in contrast to  TLKM-QPFS and IKM-QPFS approaches.
Therefore, we use {\emph QPFS with Nystr\"{o}m method} at Nystr\"{o}m sampling rate $\rho$ =0.05
for RAC dataset. The results are appended with ${ }^*$ for QPFS with Nystr\"{o}m method 
in all the tables. For RAC dataset, TLKM-QPFS and IKM-QPFS are more than two orders of magnitude 
faster than QPFS with Nystr\"{o}m on RAC dataset.

TLKM-QPFS, IKM-QPFS and IKMA-QPFS require more than an order of magnitude less memory compared to QPFS on all 
the datasets, except MNIST. On MNIST, they require about as much memory as QPFS. 
The memory required by FGM and GDM are marginally less than TLKM-QPFS, IKM-QPFS and IKMA-QPFS. 
The memory required by TLKM-QPFS, IKM-QPFS and IKMA-QPFS are comparable on all datasets.

\begin{table*}[!htb]
\begin{center}
\caption{Comparison of average execution times(in seconds).}
\label{table2-1}
\begin{tabular}{lp{0.1in}rrrrrr}
\hline
\rule{0pt}{12pt}
 {\bf Dataset}  & & {\bf QPFS} & {\bf FGM} & {\bf GDM} &  {\bf TLKM-QPFS} & {\bf IKM-QPFS} & {\bf IKMA-QPFS} \\
             
\hline \\
\quad Colon   &    &  104.90  &  2.04  &  0.46  &   {\bf 0.44}   &  1.39  &  0.76 \\
\quad SRBCT    &   &  164.42  &   -   &   -   &  11.65  &  15.22  &   {\bf 5.91}  \\
\quad Lymphoma  &     &  938.73  &   {\bf 0.21}   &  1.30  &  32.68  &  12.41  &  2.13 \\
\quad Leukemia   &    &  4864.69  &   {\bf 0.46}   &  5.91  &  17.06  &  38.35  &  38.20 \\
\quad RAC   &     & 9385.90$^{*}$  &   {\bf 0.55}   &  126.82  &  46.43  &  31.03  &  38.89 \\
\quad MNIST  &     &  161.69  &  984.16  &  268.35  &  80.37  &  60.51  &   {\bf 53.28} \\
\quad USPS   &    &  1.34  &  18.50  &  3.75  &  0.84  &   {\bf 0.80}   &  0.81

\\
\hline
\end{tabular}
\end{center}
\end{table*}

\begin{table*}[!htb]
\begin{center}
\caption{Comparison of average memory requirements(in KB).}
\label{table2-2}
\begin{tabular}{lp{0.1in}rrrrrr}
\hline
\rule{0pt}{12pt}
 {\bf Dataset}  & & {\bf QPFS} & {\bf FGM} & {\bf GDM} &  {\bf TLKM-QPFS} & {\bf IKM-QPFS} & {\bf IKMA-QPFS} \\
             
\hline \\  
\quad WDBC   &    &  544  &  1524  &  1068  &   {\bf 500}   &  524  &   524   \\
\quad Colon   &    &  84472  &  3824  &   {\bf 2472}   &  9727  &  10075  &   9884   \\
\quad SRBCT   &    &   {100418}   &   -   &   -   &  12279  &  12703  &   {\bf 11231}   \\
\quad Lymphoma   &    &  191549  &  4452  &   {\bf 2936}   &  14963  &  12504  &   10428   \\
\quad Leukemia   &    &  636103  &  10164  &   {\bf 5808}   &  13874  &  12503  &   10427   \\
\quad RAC   &    &  1456437$^{*}$  &  27284  &   {\bf 17220}   &  25879  &  21807  &   19035   \\
\quad MNIST   &    &  97076  &  253264  &   {\bf 88360}   &  97111  &  97111  &   97111   \\
\quad USPS   &    &  40138  &  14540  &   {\bf 4628}   &  33435  &  10394  &  10394

\\
\hline
\end{tabular}
\end{center}
\end{table*}

The results in tables~\ref{table2-1} and ~\ref{table2-2}, experimentally validates 
the theoretical complexities for time and memory.
%
\subsubsection{Accuracy}
\label{accuracyresults}

To compare the error rates across various 
methods, we varied the number of top features to be selected in the range from $1$ to $100$.
For RAC dataset, we varied the number of top features at an interval of $5$ 
in the range from $5$ to $100$. 
In tables~\ref{table4-1} and \ref{table4-2}, $-$ corresponding to
a method represents that the experiment was not done with that method.
From table~\ref{table4-1}, it can be observed that
our proposed IKMA-QPFS and IKM-QPFS methods achieves lowest error rates for all datasets.
In general, IKMA-QPFS and IKM-QPFS achieves lowest error rates earlier than the QPFS for 
WDBC, SRBCT, Lymphoma, Leukemia and USPS datasets and achieves lowest error rates earlier 
than FGM and GDM for all datasets except Colon dataset (tables~\ref{table4-1} 
and \ref{table4-2}). QPFS achieves lowest error rates 
earlier than FGM and GDM for WDBC, Colon, Lymphoma and Leukemia datasets.
TLKM-QPFS achieves lowest error rates earlier than the FGM for WDBC, Colon, Lymphoma 
and USPS datasets and also achieves lowest error rates earlier than the GDM for WDBC, Lymphoma 
and USPS datasets.

\begin{table*}[!htb]
\begin{center}
\caption{Table corresponding to lowest error rates (\%)}
\label{table4-1}
\begin{tabular}{lp{0.1in}rrrrrr}
\hline
\rule{0pt}{12pt}
 {\bf Dataset}  & &   {\bf QPFS}   &   {\bf FGM}   &   {\bf GDM}   &   {\bf TLKM-QPFS}   &   {\bf IKM-QPFS}   &   {\bf IKMA-QPFS} \\
\hline \\  
\quad WDBC  & &  3.28  &  3.49  &  3.26  &  3.5  &   {\bf 3.20}   &   {\bf 3.20} \\
\quad Colon &  &  12.9  &  11.29  &  16.13  &  11.29  &  9.68  &   {\bf 8.06} \\
\quad SRBCT  & &   {\bf 0.00}   &   -   &   -   &   {\bf 0.00}   &   {\bf 0.00}   &   {\bf 0.00}  \\
\quad Lymphoma &  &   {\bf 0.00}   &  2.22  &  6.67  &   {\bf 0.00}   &   {\bf 0.00}    &   {\bf 0.00} \\
\quad Leukemia &  &  2.78  &  11.11  &  16.67  &  1.39  &   {\bf 0.00}   &   {\bf 0.00} \\
\quad RAC  & &   {\bf 0.00$^{*}$}   &  6.06  &   {\bf 0.00}   &   {\bf 0.00}   &   {\bf 0.00}    &   {\bf 0.00} \\
\quad MNIST &  &  4.13  &  4.69  &  4.99  &  3.83  &  3.48  &   {\bf 3.43} \\
\quad USPS  & &  9.02  &  9.86  &  10.1  &  9.49  &   {\bf 8.27}  &   {\bf 8.27} \\

\hline
\end{tabular}
\end{center}
\end{table*}

\begin{table*}[!htb]
\begin{center}
\caption{Table for number of features corresponding to lowest error rates (\%)}
\label{table4-2}
\begin{tabular}{lp{0.1in}rrrrrr}
\hline
\rule{0pt}{12pt}
 {\bf Dataset}   &    &   {\bf QPFS}   &   {\bf FGM}   &   {\bf GDM}   &   {\bf TLKM-QPFS}   &   {\bf IKM-QPFS}   &   {\bf IKMA-QPFS} \\
\hline \\
\quad WDBC   &    &  21  &  29  &  25  &  14  &  10  &   10  \\
\quad Colon   &    &  5  &  44  &  6  &  14  &  35  &   24  \\
\quad SRBCT   &    &  11  &   -   &   -   &  9  &  8  &   8  \\
\quad Lymphoma   &    &  17  &  47  &  100  &  8  &  7  &   6  \\
\quad Leukemia   &    &  24  &  37  &  38  &  39  &  34  &   34  \\
\quad RAC   &    &  10$^{*}$  &  7  &  4  &  9  &  7  &   5  \\
\quad MNIST   &    &  95  &  20  &  75  &  85  &  99  &   99  \\
\quad USPS   &    &  93  &  54  &  64  &  34  &  67  &   67  \\

\hline
\end{tabular}
\end{center}
\end{table*}

Figure~\ref{fig2} plots the error rates for each datasets as 
the number of top selected features is varied from $1$ to $100$.
The baseline here represents the accuracy obtained when all the features are used for
classification and k-means-baseline represents the accuracy obtained when all the representative 
features (after two-level k-means) are used for classification.
It is evident from figures~\ref{fig2a}-\ref{fig2f}
%
%
that error rates achieved by TLKM-QPFS, IKM-QPFS and IKMA-QPFS 
methods are improved over QPFS, FGM and GDM for each of the datasets. 
In all the datasets, TLKM-QPFS and IKM-QPFS achieve lower error rates with a less 
number of top selected features than QPFS. Usually, IKM-QPFS and IKMA-QPFS achieves
lower error rates early than the TLKM-QPFS. In figure~\ref{fig2} and figure~\ref{fig3}, plots of 
IKM-QPFS and IKMA-QPFS significantly overlaps. 

%
As expected, the error rates come down as relevant features are added to the set. Once the relevant set
has been added, any more additional (irrelevant) features lead to loss in accuracy. 
Tables~\ref{table3-1} and~\ref{table3-2} present the average test set error rates for each of the methods
for top ranked $k$ features ($k$ being 10, 20, 30,50, 100), where top
$k$ features are chosen as output by the respective feature selection method.
On all the datasets, IKMA-QPFS performs significantly better than QPFS, FGM and GDM
 at all the values of top $k$ features selected. Further, error rates for TLKM-QPFS, IKM-QPFS
 and IKMA-QPFS are comparable on all datasets. 
This is particularly
evident early on i.e. for a smaller number of top-$k$ features. This points to
the fact that IKM-QPFS and IKMA-QPFS are able to rank the relevant set of features right 
at the top. 

TLKM-QPFS performs better than QPFS in all the cases (dataset and number
of top $k$ feature combination), except on Colon data at 
$k=10$~\footnote{it performs marginally worse in couple of cases ($k=10$, $k=30$)
for SRBCT}. Among the two proposed approaches (TLKM-QPFS, IKM-QPFS and IKMA-QPFS),
both IKM-QPFS and IKMA-QPFS are clear winner in terms of the accuracy.

\subsection{Summary}
It is clearly evident from above results that our all the three proposed approaches
 for feature selection, TLKM-QPFS, IKM-QPFS and IKMA-QPFS, give significant gains in
 computational requirements (both time and memory), even while improving the overall accuracy
in all cases when compared with QPFS and significantly low error rates when compared with
FGM and GDM. Especially, our proposed approaches help reach the relevant set of features early 
on, which is a very important property of a good feature selection method. The 
computational requirements of TLKM-QPFS, IKM-QPFS and IKMA-QPFS are similar to each
other. On the large microarray dataset our proposed approaches are faster
than the FGM and GDM. As for performance, IKMA-QPFS is a clear winner among the three variants

In tables~\ref{table3-1} and ~\ref{table3-2}, TLKM, IKM and IKMA represent 
TLKM-QPFS, IKM-QPFS and IKMA-QPFS, respectively.
\setlength{\tabcolsep}{1.8pt}
\begin{table}[!htb]
\caption{Error rates (\%) for bioinformatics datasets by each methods} 
\label{table3-1}
\centering
\begin{tabular}{llrrrrr}

\hline
&&\multicolumn{5}{c}{\bf k (number of top features)} \\[-1ex]
\raisebox{1.5ex}{\bf Dataset} &\raisebox{1.5ex}{}
& {\bf 10} & {\bf 20} & {\bf 30} & {\bf 50} & {\bf 100} 
\\ [0.5ex]
\hline
& QPFS& 6.07 & {\bf 3.41} & 3.49 & - & -  \\[0.2ex]
& FGM & 3.61 & 3.64 & 3.49 & - & -  \\[0.2ex]
& GDM & 4.54 & 3.72 & 3.36 & - & -  \\[-1ex]
\raisebox{1.5ex}{WDBC} 
& TLKM & 4.43 & 3.72 & - & - & -  \\[0.2ex]
& IKM & {\bf 3.20} & 3.43 & - & - & -  \\[0.2ex]
& IKMA & {\bf 3.20} & 3.43 & - & - & -  \\[0.2ex]
\hline

& QPFS&{\bf 14.52} & 24.19 & 19.36 & 20.97 & 24.19  \\[0.2ex]
& FGM & 19.36 & 16.13 & 16.13 & 17.74 & {\bf 14.52}  \\[0.2ex]
& GDM & 22.58 & 30.65 &  30.65 & 25.81 &  24.19  \\[-1ex]
\raisebox{1.5ex}{Colon} 
& TLKM & 20.97 & 14.52 & {\bf 14.52} & 17.74 & 17.74  \\[0.2ex]
& IKM & 17.74 & 22.58 & {\bf 14.52} & {\bf 14.52} & 19.35  \\[0.2ex]
& IKMA & 16.13 & {\bf 12.90} & 16.13 & 22.58 & 20.97  \\[0.2ex]
\hline

& QPFS&3.18 & 0.00 & 0.00 & 0.00 & 0.00  \\[0.2ex]
& FGM & - & - & - & - & -  \\[0.2ex]
& GDM & - & - & - & - & -  \\[-1ex]
\raisebox{1.5ex}{SRBCT} 
& TLKM & 1.59 & 0.00 & 1.59 & 0.00 & 0.00  \\[0.2ex]
& IKM & 0.00 & 0.00 & 0.00 & 0.00 & 0.00  \\[0.2ex]
& IKMA & 0.00 & 0.00 & 0.00 & 0.00 & 1.59  \\[0.2ex]
\hline

& QPFS& 8.89 & 2.22 & 0.00 & 0.00 & 0.00  \\[0.2ex]
& FGM & 6.67 & 8.89 & 8.89 & 8.89 & 4.44  \\[0.2ex]
& GDM & 17.78 & 28.89 & 33.33 & 28.89 & 6.67  \\[-1ex]
\raisebox{1.5ex}{Lymphoma} 
& TLKM & 4.44 & 0.00 & 0.00 &0.00 & 0.00  \\[0.2ex]
& IKM & 8.89 & 0.00 & 0.00 & 0.00 & 0.00  \\[0.2ex]
& IKMA & 0.00 & 0.00 & 0.00 & 0.00 & 4.44  \\[0.2ex]
\hline

& QPFS&{\bf 13.89} & {\bf 9.72} & 5.56 & 6.94 & 5.56  \\[0.2ex]
& FGM & 20.83 & 19.44 & 19.44 & 13.85 & 13.89  \\[0.2ex]
& GDM & 18.06 & 25.00 & 31.94 & 18.06 & 27.78  \\[-1ex]
\raisebox{1.5ex}{Leukemia} 
& TLKM & 18.06 & 15.28 & {\bf 4.17} & 5.56 &5.56  \\[0.2ex]
& IKM & {\bf 13.89} & 13.89 & 6.94 & {\bf 4.17} & {\bf 4.17}  \\[0.2ex]
& IKMA & {\bf 13.89} & 13.89 & 6.94 & {\bf 4.17} & {\bf 4.17}  \\[0.2ex]
\hline

& QPFS+Nys& 0.00 & 0.00 & 3.03 & 0.00 & 0.00  \\[0.2ex]
& FGM & 9.09 & 9.09 & 15.15 & 15.15 & 18.18  \\[0.2ex]
& GDM & 3.03 & 9.09 & 9.09 & 24.24 & 12.5  \\[-1ex]
\raisebox{1.5ex}{RAC} 
& TLKM & 0.00 & 0.00 & 0.00 & 0.00 & 0.00  \\[0.2ex]
& IKM & 0.00 & 0.00 & 0.00 & 0.00 & 0.00 \\[0.2ex]
& IKMA & 0.00 & 3.03 & 9.09 & 9.09 & -  \\[0.2ex]

\\
\hline

\end{tabular}
\end{table}

 \setlength{\tabcolsep}{1.8pt}
\begin{table}[!htb]
\caption{Error rates (\%) for vision datasets by each methods} 
\label{table3-2}
\centering
\begin{tabular}{llrrrrr}

\hline
&&\multicolumn{5}{c}{\bf k (number of top features)} \\[-1ex]
\raisebox{1.5ex}{\bf Dataset} &\raisebox{1.5ex}{}
& {\bf 10} & {\bf 20} & {\bf 30} & {\bf 50} & {\bf 100} 
\\ [0.5ex]
\hline
& QPFS& 15.83 & 9.47 & 6.25 & 5.40 & 4.83  \\[0.2ex]
& FGM & 6.20 & {\bf 4.69} & 8.42 & 9.32 & 14.77  \\[0.2ex]
& GDM & {\bf 5.79} & 5.79 & 5.79 & 5.79 & 5.85  \\[-1ex]
\raisebox{1.5ex}{MNIST} 
& TLKM & 10.18 & 6.40 & 5.99 & 4.89 & 4.03  \\[0.2ex]
& IKM & 15.73 & 5.50 & 4.99& {\bf 4.3} & 3.53 \\[0.2ex]
& IKMA & 15.32 & 6.15 & {\bf 4.33} & 4.74 & {\bf 3.48}  \\[0.2ex]
\hline

& QPFS & 16.41 & 13.13 & 12.06 & 11.65 & {\bf 9.20}  \\[0.2ex]
& FGM & 12.82 & 10.96 & 11.13 & 9.89 & 10.30  \\[0.2ex]
& GDM & {\bf 10.37} & {\bf 10.33} & 10.45 & 10.39 & 11.23  \\[-1ex]
\raisebox{1.5ex}{USPS} 
& TLKM & 16.51 & 11.76 & {\bf 9.96} & 9.83 & 9.89  \\[0.2ex]
& IKM & 18.49 & 15.99 & 11.76& {\bf 9.20} & 9.32 \\[0.2ex]
& IKMA& 18.48 & 15.98 & 11.76& {\bf 9.20} & 9.32 \\[0.2ex]
\\
\hline

\end{tabular}
\end{table}



\begin{figure*}[!htb]
\centering
  \subfigure[WDBC Dataset]{\label{fig2a}\input{texfigicdm/wdbc.tex}} \hspace{0.5cm}         
  \subfigure[Colon Dataset]{\label{fig2b}\input{texfigicdm/colon.tex}}\\
  \subfigure[SRBCT Dataset]{\label{fig2c}\input{texfigicdm/srbct.tex}} \hspace{0.5cm}      
  \subfigure[Lymphoma Dataset]{\label{fig2d}\input{texfigicdm/lymphoma.tex}} \\ 
  \subfigure[Leukemia Dataset]{\label{fig2e}\input{texfigicdm/leukemia.tex}} \hspace{0.5cm}         
  \subfigure[RAC Dataset]{\label{fig2f}\input{texfigicdm/rac.tex}} 
\caption{Plots of Error rates for each methods with varying number of top $k$(1-100) features for bioinformatics datesets}
\label{fig2}
\end{figure*}

\begin{figure*}[!htb]
\centering
\subfigure[MNIST Dataset]{\label{fig3a}\input{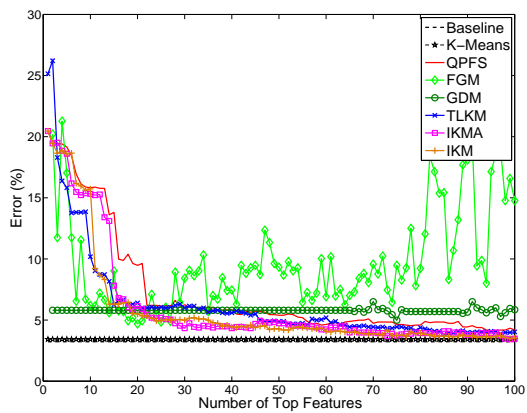}} \hspace{0.5cm}         
  \subfigure[USPS Dataset]{\label{fig3b}\input{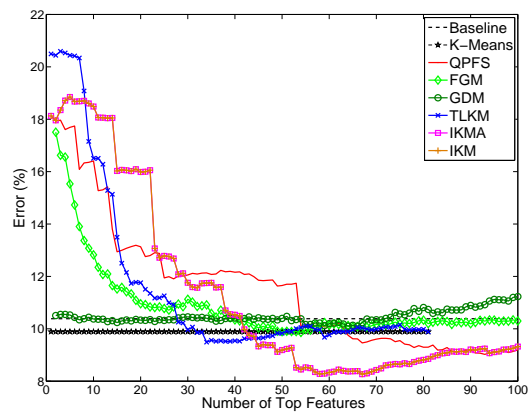}} 
\caption{Plots of Error rates for each methods with varying number of top $k$(1-100) features for vision datasets}
\label{fig3}
\end{figure*}
\section{Conclusion}
\label{conc}
In this paper, we proposed an approach for integrating k-means based clustering
with Quadratic Programming Feature Selection (QPFS). The key idea involved
using k-means to cluster together redundant sets of features. Only one representative
from each cluster needed to be considered during the QPFS run for feature selection,
reducing the complexity of QPFS from cubic in number of features to cubic in
number of clusters (which is much smaller). We presented two variations of our 
approach. TLKM-QPFS used two level k-means to identify a set of representative 
features followed by a run of QPFS. In the more sophisticated variant, IKMA-QPFS,
we interleaved the steps of k-means with QPFS, leading to a very fine grained
selection of relevant features. Extensive evaluation on eight publicly available
datasets showed the superior performance of our approach relative to existing
state of the art feature selection methods. 

One of the key directions for future work involves providing a generic framework 
for integrating a given clustering algorithm with a filter based feature selection 
method. Other direction includes extending our approach to sparse representations 
to deal with data in very high dimensions (millions of features, such as in vision). 
A third direction deals with coming up with a parallel formulation of our proposed 
approach.

\section*{Acknowledgment}
The authors would like to thank Dr. Parag Singla, Department 
of Computer Science \& Engineering, Indian Institute of Technology,
Delhi, India for his valuable suggestions and support.
%
%
%
%
%
\bibliography{icdm13}
\bibliographystyle{elsarticle-num}

\end{document}